\documentclass[lettersize,journal]{IEEEtran}
\usepackage{amsmath,amsfonts}
\usepackage{algorithm}
\usepackage{algpseudocode}
\usepackage{bm} 
\usepackage{array}
\usepackage{textcomp}
\usepackage{stfloats}
\usepackage{url}
\usepackage{verbatim}
\usepackage{graphicx}
\usepackage{cite}
\usepackage{subcaption}
\usepackage[font=small]{caption}
\usepackage{booktabs}   
\usepackage{multirow}   
\usepackage[linkcolor=black,citecolor=black,urlcolor=black,colorlinks=TRUE]{hyperref}
\usepackage{tabularx}

\graphicspath{{./FIGURES/}}
\DeclareGraphicsExtensions{.png,.jpg,.eps,.pdf}

\begin{document}

\title{\LARGE \bf
	High-Speed Vision-Based Flight in Clutter with Safety-Shielded Reinforcement Learning
}

 \author{Jiarui Zhang\textsuperscript{1,2$\dagger$},
	    Chengyong Lei\textsuperscript{1,2$\dagger$},
	    Chengjiang Dai\textsuperscript{1,2},
        Kenghou Hoi\textsuperscript{1,2},
	    Lijie Wang\textsuperscript{1,2},\\
	    Zhichao Han\textsuperscript{1,2*},
	    and Fei Gao\textsuperscript{1,2*}
	 	\thanks{\textsuperscript{$\dagger$}These authors contributed equally to this work.}
	 	\thanks{\emph{\textsuperscript{*}Corresponding authors: Zhichao Han and Fei Gao}}
	 	\thanks{\textsuperscript{1}Institute of Cyber-Systems and Control, College of Control Science and Engineering, Zhejiang University, Hangzhou 310027, China}
	 	\thanks{\textsuperscript{2}Differential Robotics, Hangzhou 311121, China.}
	 	\thanks{E-mail:{\tt\small \{jrzzz, fgaoaa\}@zju.edu.cn}}
	 }


\maketitle
\thispagestyle{empty}
\pagestyle{empty}

\begin{abstract}
Quadrotor unmanned aerial vehicles (UAVs) are increasingly deployed in complex missions that demand reliable autonomous navigation and robust obstacle avoidance. However, traditional modular pipelines often incur cumulative latency, whereas purely reinforcement learning (RL) approaches typically provide limited formal safety guarantees. To bridge this gap, we propose an end-to-end RL framework augmented with model-based safety mechanisms. We incorporate physical priors in both training and deployment. During training, we design a physics-informed reward structure that provides global navigational guidance. During deployment, we integrate a real-time safety filter that projects the policy outputs onto a provably safe set to enforce strict collision-avoidance constraints. This hybrid architecture reconciles high-speed flight with robust safety assurances. Benchmark evaluations demonstrate that our method outperforms both traditional planners and recent end-to-end obstacle avoidance approaches based on differentiable physics. Extensive experiments demonstrate strong generalization, enabling reliable high-speed navigation in dense clutter and challenging outdoor forest environments at velocities up to $7.5 \text{ m/s}$.
\end{abstract}


\section{Introduction}
\IEEEPARstart{I}{n} recent years, quadrotor Unmanned Aerial Vehicles (UAVs) have been widely adopted in logistics, search and rescue, and exploration missions, owing to their exceptional maneuverability and agility~\cite{Zhou2021FUEL, xu2023vision}. However, navigation in these unstructured and dynamic environments requires the system to perceive sudden obstacles and execute evasive maneuvers within short time. This creates a fundamental conflict between the demand for high-speed flight and the computational latency required to guarantee rigorous collision avoidance.

Traditional navigation pipelines decompose autonomy into sequential perception, planning, and control modules, achieving reliable performance through explicit algorithmic structures and well-established safety margins~\cite{zhou2019robust, gao2020teach, zhou2022swarm}. However, this modular design inevitably incurs cumulative latency, as perception updates, global replanning, and trajectory optimization are repeatedly executed in a closed loop, leading to delayed reactions in rapidly changing environments~\cite{wu2024whole}. Moreover, planning in densely cluttered spaces often requires high-dimensional search or optimization, causing computational costs to scale poorly with environmental complexity and limiting real-time performance at high flight speeds.

In contrast, learning-based navigation has emerged as an attractive alternative by directly mapping sensory observations to control commands~\cite{kahn2021badgr,nguyen2022motion, song2022learning, yu2024mavrl, zhang2025learning}. Such end-to-end policies significantly reduce processing latency and have demonstrated superior agility in high-speed flight scenarios compared to modular pipelines~\cite{loquercio2021learning}. Despite these advantages, purely learning-based approaches typically lack theoretical safety guarantees and are vulnerable to out-of-distribution generalization failures, which fundamentally constrain their reliability in safety-critical applications.

To address the lack of formal safety guarantees in purely end-to-end navigation policies while preserving their agility in complex environments, we propose a hybrid navigation framework that tightly couples reinforcement learning with model-based safety mechanisms. The key idea is to exploit the low-latency and expressive power of end-to-end policies for high-speed maneuvering, while enforcing safety through explicit physical constraints. During training, we incorporate model-based priors into the learning process by shaping the reward with global path guidance and safety-related constraints, encouraging the policy to acquire navigation behaviors that are both efficient and risk-aware. During deployment, we further introduce a real-time safety correction module based on high-order control barrier functions, which projects policy outputs onto a provably safe set to guarantee collision avoidance under unforeseen disturbances and distribution shifts. In addition, we employ robust sim-to-real transfer strategies to ensure consistent performance on real quadrotor platforms.

Our main contributions are summarized as follows:

\begin{itemize}
	\item [1)] A training-time safety-aware navigation framework that integrates global path structure and local collision avoidance into end-to-end reinforcement learning. The proposed approach alleviates local minima in Euclidean distance objectives and enables more foresighted navigation behaviors in cluttered environments.
	\item [2)] A safety enforcement mechanism based on control barrier functions, which guarantees collision avoidance by correcting policy outputs in real time. This mechanism complements learning-based agility with formal safety constraints, enabling high-speed flight without sacrificing reliability under distribution shifts.
	\item [3)] A comprehensive experimental validation across simulation and real-world platforms. Extensive benchmarks demonstrate superior performance over both traditional planners and learning-based baselines, while real-world experiments validate robust sim-to-real transfer and agile navigation in densely cluttered environments.
\end{itemize}

\section{Related Work}
\subsection{Modular Navigation for UAV}
Traditional optimization-based navigation methods typically rely on explicit 3D environmental representations constructed from depth measurements and manually derived safety constraints~\cite{zhou2019robust, scaramuzza2014vision, zhou2020ego}. Hard-constrained approaches, for example, decouple planning into extracting a Safe Flight Corridor(SFC)-represented by cubes, spheres, or polyhedrons and subsequently generating trajectories via convex optimization~\cite{liu2017sfc}. While ensuring global optimality within the corridor, these methods often depend on heuristic time allocation strategies, which can degrade trajectory quality and lead to conservative dynamic constraints that limit agility. 

Conversely, soft-constrained methods using Euclidean Signed Distance Fields (ESDF) jointly optimize planning and control, leveraging gradient information to converge upon feasible trajectories within continuous space.  However, maintaining real-time ESDF imposes substantial computational overhead on onboard processors due to the expensive calculation of line integral calculations, necessitating a difficult trade-off between mapping precision and update frequency. To mitigate this, Ego-Planner~\cite{zhou2020ego} circumvents explicit ESDF construction by iteratively generating safe guidance paths to provide obstacle avoidance gradients. Nevertheless, such methods lack theoretical convergence guarantees and remain susceptible to entrapment in unsafe local minima within complex, non-convex environments. Furthermore, the cascading architecture of these multi-module systems introduces significant parameter complexity and cumulative latency, leading to state estimation instability during high-speed flight and limiting robustness in dynamic scenarios.

\subsection{Learning-Based Methods}
Recent studies underscore the immense potential of learning-based methodologies, which obviate the need for dense mapping or explicit environmental modeling. By directly mapping sensory observations to control commands, these approaches circumvent the cumulative latency and computational burden associated with modular trajectory optimization, demonstrating impressive maneuverability~\cite{han2025reactive, xu2025flying,wu2026precise}. NavRL~\cite{xu2025navrl} introduces a PPO-based framework that enables safe and reactive navigation in dynamic environments with moving obstacles, achieving low collision rates through careful reward design and curriculum training. Similarly, approaches such as ~\cite{yu2024mavrl} and ~\cite{wu2024clutter} proposes an RL pipeline that dynamically adapts flight speed based on environmental clutter, striking an effective balance between success rate and agility in unknown cluttered spaces.  Furthermore, emerging research has leveraged differentiable simulators and optimization-embedded networks to refine controller performance. hybrid approaches~\cite{han2025dynamically} embed traditional trajectory optimization within neural networks to generate dynamically feasible trajectories directly from depth inputs, bridging perception and planning while ensuring kinematic constraints without explicit mapping.

While these learning-based approaches have propelled UAV agility to new heights, their inherent lack of theoretical completeness often compromises reliability in safety-critical scenarios. 
Hybrid architectures integrating Control Barrier Functions (CBFs) have emerged as a prominent approach to address this limitation. For example, the RL-CBF framework proposed by Cheng et al. couples model-free RL algorithms with model-based CBF controllers to ensure safety throughout the training process~\cite{cheng2019end}. To overcome the challenges of integrating hard constraints into gradient-based learning, Emam et al. introduced a differentiable Robust Control Barrier Function safety layer embedded within a Soft Actor-Critic framework~\cite{emam2022safe}. Concurrently, recent approaches have demonstrated the effectiveness of incorporating global path structures as privileged information during training to prevent policies from becoming trapped in non-convex obstacles. For instance, Lee et al. utilize Time-of-Arrival (ToA) maps to define explicit velocity setpoint gradients, effectively guiding the policy through highly cluttered spaces and dead ends~\cite{lee2025quadrotor}. While these prior works elegantly embed analytical models into neural networks, we propose a distinct, position-based reward within a PPO framework~\cite{ppo}. By avoiding strict velocity tracking, this hybrid strategy bridges agility and safety, endowing the policy with enhanced robustness and global reachability without overly constraining maneuverability.

\section{Methodology}

\subsection{Problem Formulation}
We formulate the real-time interaction between the simulated UAV and its complex, dynamic surroundings as a Markov Decision Process (MDP). Formally, the MDP is defined as the tuple $(\mathcal{S}, \mathcal{A}, \mathcal{P}, \mathcal{R}, \gamma)$, where $\mathcal{S}$ denotes the state space, $\mathcal{A}$ the action space, $\mathcal{P}: \mathcal{S} \times \mathcal{A} \times \mathcal{S} \rightarrow [0, 1]$ the state-transition probability, $\mathcal{R}: \mathcal{S} \times \mathcal{A} \rightarrow \mathbb{R}$ the reward function, and $\gamma \in [0, 1)$ the discount factor. 
The actor-critic network maps directly from $\mathcal{S}$ to $\mathcal{A}$ rather than a multi-stage or explicit curriculum learning approach. The simulation step size is set to $0.01$\,s, so both the policy output and the control execution operate at $100$\,Hz.    
Detailed descriptions of the observation space, action space, and reward design are provided in the subsequent subsections~\ref{sec:obs}, ~\ref{sec:act} and ~\ref{sec:reward}.

\subsubsection{Observation Space}
\label{sec:obs}
The training loop initiates by acquiring the current observation of the UAV state. The observation input to our policy network is structured as a multi-modal composite vector $\mathbf{o}_t$, integrating both exteroceptive environmental perception and the proprioceptive state of the UAV. Specifically, the observation is defined as:
\begin{equation}
	\mathbf{o}_t = \left[ \mathbf{D}_t, \mathbf{v}_t, \mathbf{R}_t, \mathbf{a}_{t-1}^{\pi}, \mathbf{d}^{xy}_t, z^{\text{self}}_{t}, z^{\text{goal}}_{t} \right],
\end{equation}
where the components are detailed as follows:

\begin{itemize}
	\item $\mathbf{D}_t \in \mathbb{R}^{100 \times 60}$: the depth image captured by the onboard camera.
	
	\item $\mathbf{v}_t \in \mathbb{R}^3$: the linear velocity vector expressed in the body coordinate frame.
	
	\item $\mathbf{R}_t \in SO(3)$: the rotation matrix representing the current attitude of the UAV.
	
	\item $\mathbf{a}_{t-1}^{\pi}$: the action executed at the previous time step, comprising roll, pitch, yaw, and total thrust. This historical action serves as a reference to ensure control smoothness.
	
	\item $\mathbf{d}^{xy}_{t} = [\Delta x, \Delta y]^\top$: the normalized horizontal position difference relative to the target.
	
	\item $z^{\text{self}}_{t}$ and $z^{\text{goal}}_{t}$: the current altitude of the UAV and the target altitude, respectively.
\end{itemize}

\begin{figure*}[t]
	\centering
	\includegraphics[width=1\linewidth]{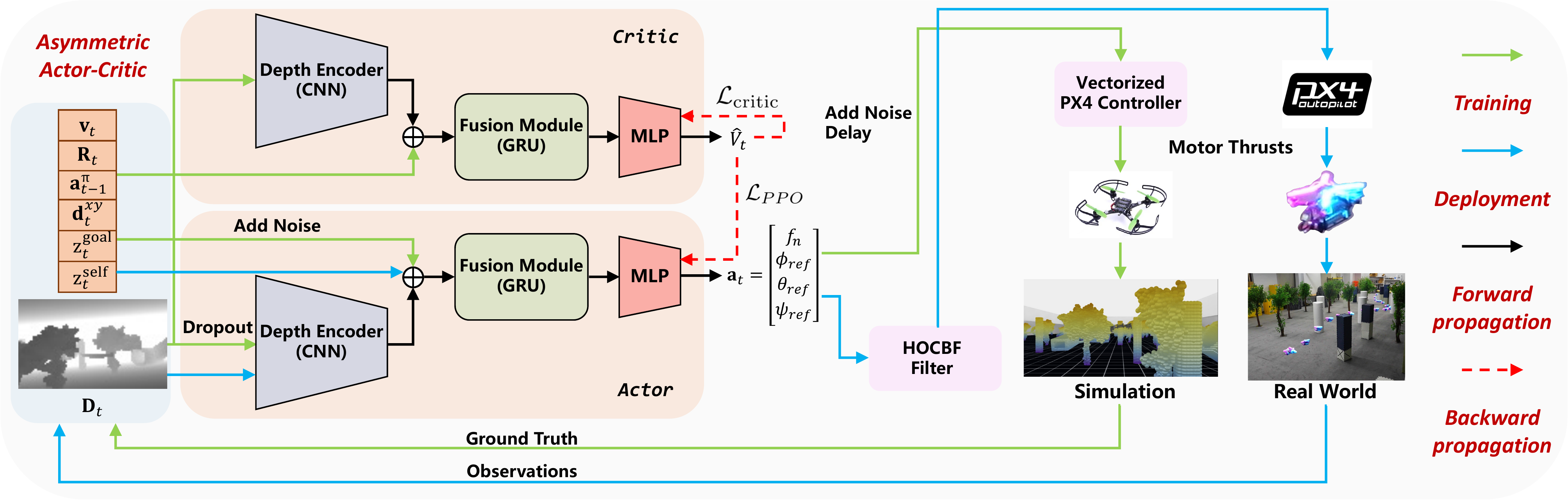} 
	\captionsetup{font={small}}
	\caption{\textbf{Network architecture and control pipeline.}
		An asymmetric actor-critic policy fuses a depth image (CNN) and proprioceptive states via a GRU.
		The actor outputs attitude references and normalized thrust, which are tracked by a PX4 controller.
		During training, domain randomization (dropout/noise/delay) improves robustness; during deployment, a real-time HOCBF filter refines the raw command to enforce safety constraints.} 
	
	\vspace{-4pt}
	\label{fig:network}
\end{figure*}

\subsubsection{Action Space}
\label{sec:act}
The policy network processes the aforementioned observations and outputs a four-dimensional action vector $\mathbf{a}_t^{\pi} \in \mathbb{R}^4$, defined as:
\begin{equation}
	\mathbf{a}_t^{\pi} = [f_n, \phi_{ref}, \theta_{ref}, \psi_{ref}]^\top,
\end{equation}
this vector consists of the mass-normalized collective thrust $f_n$ and the desired attitude angles (roll $\phi_{ref}$, pitch $\theta_{ref}$, and yaw $\psi_{ref}$). The control commands are updated at a fixed frequency of 100 Hz to guarantee real-time responsiveness. Subsequently, the low-level onboard flight controller solves for the required rotational speeds of individual motors to achieve precise tracking of these high-level commands.

To ensure operational safety, real-world UAVs typically enforce bounded attitude commands to prevent aggressive maneuvers that may lead to loss of control. Accordingly, instead of issuing low-level control inputs such as direct motor speeds, we adopt attitude-level control commands as the policy output. This higher-level control abstraction significantly reduces the sim-to-real gap by aligning the action space in simulation with practical flight controllers used on physical platforms, thereby improving the robustness of sim-to-real transfer~\cite{song2023reaching}.

\subsubsection{Reward Function Design}
\label{sec:reward}

In this work, the overall reward $r_t$ is defined as a weighted sum of distinct components:
\begin{equation}
	\begin{aligned}
		r_t =\;& 
		\alpha_1 r_{\mathrm{navigation}}
		+ \alpha_2 r_{\mathrm{safety}}
		+ \alpha_3 r_{\mathrm{smooth}} \\[4pt]
		&+ \alpha_4 r_{\mathrm{collision}}
		+ \alpha_5 r_{\mathrm{success}} \\[4pt]
		&+ \alpha_6 r_{\mathrm{speed}}
		+ \alpha_7 r_{\mathrm{height}} .
	\end{aligned}
\end{equation}
Specifically, the total reward is composed of navigation guidance, safety regulation, control regularization, and terminal signals. The model-based components $r_{\mathrm{navigation}}$ and $r_{\mathrm{safety}}$, detailed in Section \ref{sec:model-based-reward-shaping}, provide global path guidance and safety-aware shaping during training. To ensure executable and stable control behavior, the smoothness term $r_{\mathrm{smooth}}$ penalizes both large control magnitudes and abrupt changes between consecutive actions, encouraging dynamically feasible and smooth trajectories. Terminal rewards $r_{\mathrm{collision}}$ and $r_{\mathrm{success}}$ are applied upon collision events and successful task completion, respectively, providing clear episodic feedback. In addition, auxiliary terms. $r_{\mathrm{speed}}$ and $r_{\mathrm{height}}$ softly constrain the UAV’s velocity and altitude within predefined safety limits, preventing unrealistic or unsafe behaviors during exploration. The overall reward is formulated as a weighted sum, where $\alpha_i$ denotes the coefficient of the $i$-th term. These weights are empirically tuned to balance safety, agility, and task completion during training.

\subsection{Network Architecture}

We employ the Proximal Policy Optimization (PPO) algorithm within an asymmetric actor-critic framework. This architecture leverages a privileged learning paradigm: while the critic network is trained on noise-free ground truth states to facilitate stable value estimation, the actor network operates solely on noisy observations to ensure robust deployment under real-world uncertainties.

The complete network structure is illustrated in Fig.~\ref{fig:network}.To effectively process multimodal inputs, the actor utilizes a heterogeneous neural architecture. The visual stream processes a $100 \times 60$ depth image via a Convolutional Neural Network (CNN) composed of sequential convolutional, activation (ReLU), and pooling layers, extracting a compact visual feature vector $f_{\text{CNN}}$ that encodes spatial geometry. We directly concatenate this visual embedding with the raw proprioceptive observations, which consist of the quadrotor’s linear velocity, rotation matrix, previous action, and goal-relative displacement. This unified vector is subsequently fed into a Gated Recurrent Unit (GRU). The GRU maintains a temporal hidden state $h_t$ to aggregate historical context, enabling the policy to mitigate partial observability and smooth sensor noise over time. Finally, the GRU output is passed to a Multi-Layer Perceptron (MLP) head, which predicts the Gaussian distribution parameters for the continuous action command $\mathbf{a}_t$.

\subsection{Model-based Reward Shaping}
\label{sec:model-based-reward-shaping}
We begin by formulating the navigation reward $r_{\mathrm{navigation}}$. Conventional distance-based rewards typically rely on Euclidean distance to shape the policy. Although this offers dense feedback, it often leads to convergence failures in non-convex environments. The agent, driven by the greedy minimization of Euclidean distance, is prone to the local minima trap, failing to navigate around complex obstacles. To overcome this, we propose a geodesic-guided reward mechanism. Instead of Euclidean distance, we utilize a precomputed cost map derived from Dijkstra algorithm, which represents the true shortest-path distance considering obstacles. This map serves as a dense potential field during training. By applying trilinear interpolation, we extract continuous cost values $\mathbf{V}_t$ at the real-time coordinates of the UAV. The reward is then formulated based on the progress $\bm{\delta}$ along this potential field. This approach provides the agent with global geometric awareness, effectively decomposing the long-horizon task into accessible local subgoals and guiding the UAV out of local minima.

Let $\Phi_g$ denote the precomputed geodesic distance field (via Dijkstra) for a given scene $g$. The navigation reward $r_{\mathrm{navigation}, t}$ at time step $t$ is defined as the clipped progress made along this potential field:
\begin{equation}
	\resizebox{0.91\hsize}{!}{$
		r_{\mathrm{navigation}, t} = \lambda \cdot \text{clamp}( \underbrace{\text{Interp}(\Phi_g, \mathbf{p}_{t-1})}_{\mathbf{V}_{t-1}} - \underbrace{\text{Interp}(\Phi_g, \mathbf{p}_t)}_{\mathbf{V}_t}, -C, C ),
		$}
\end{equation}
where $\text{Interp}(\cdot)$ represents the trilinear interpolation function that maps the continuous UAV position $\mathbf{p}$ to the discrete Dijkstra cost grid, $\lambda$ is a scaling factor, and the $\text{clamp}$ function ensures numerical stability by bounding the reward within $[-C, C]$.

To enhance flight safety, we incorporate a shaping reward based on Control Barrier Function~\cite{ames2016control,ames2019control, cheng2019end}. We first construct a ESDF from the obstacle map, where $d(\mathbf{x})$ denotes the minimum distance from position $\mathbf{x}$ to the nearest obstacle. The barrier function is defined as $b(\mathbf{x}) = d(\mathbf{x}) - d_{\text{safe}}$, where $d_{\text{safe}}$ is a prescribed safety margin. According to CBF theory, safety is guaranteed if the forward-invariance condition $\dot{b}(\mathbf{x}) + \gamma b(\mathbf{x}) \ge 0$ is satisfied. Instead of enforcing this constraint explicitly, we encode the violation of this condition into the reward function. Specifically, at each timestep $t$, we compute the time derivative of the barrier function as $\dot{b}(\mathbf{x}_t) = \nabla d(\mathbf{x}_t) \cdot \mathbf{v}_t$, utilizing the gradient of the ESDF and the current velocity vector of the UAV. The safety reward $r_{\text{safety}}$ is then formulated as:
\begin{equation}
	r_{\text{safety}} = \text{clip}  \left( \dot{b}(\mathbf{x}_t) + \gamma b(\mathbf{x}_t) ,  \delta_{\text{min}} , 0 \right ),
\end{equation}
where $\gamma$ is the CBF coefficient. We impose a clipping threshold $\delta_{\text{min}} = -2.0$ on the safety reward. This lower bound prevents unbounded penalties from destabilizing the value function approximation and causing gradient explosion, thereby ensuring numerical stability during training. This mechanism encourages the UAV to proactively align its velocity with the ESDF gradient to satisfy the barrier condition, effectively anticipating and avoiding collision risks.

Although we utilize privileged map information to construct $r_{navigation}$ and $r_{safety}$ to enhance the policy's navigation capability and safety, both of these rewards are strictly used only during the training phase and not during deployment.

\subsection{HOCBF-based Correction Loop}

Since reward shaping only acts as a soft incentive rather than a hard constraint, the agent can still suffer severe failures. To provide theoretical safety guarantees during real-world deployment, we integrate a safety filter based on High-Order Control Barrier Functions (HOCBF)~\cite{xiao2021hocbf}. 
This optimization layer minimally deviates from the raw reference acceleration $  \mathbf{a}_{\text{raw}}  $which can be derived from the policy output $  \mathbf{a}_t^\pi  $ to strictly satisfy the safety constraints.  
We formulate this as a Quadratic Program (QP):
\begin{equation} 
	\mathbf{a}^* = \underset{\mathbf{a}} {\arg\min}\ \frac{1}{2} | \mathbf{a} - \mathbf{a}_\text{raw} |^2 \quad \text{s.t.} \quad \mathcal{C}(\mathbf{r}_t, \mathbf{a}) \geq 0 .
\end{equation}
For an obstacle $i$, we define the candidate barrier function $B_i(\mathbf{r}_t) = \| \mathbf{r}_t \|^2 - r_{safe}^2$, where $\mathbf{r}_t$ denotes the relative position vector extending from the UAV to the nearest obstacle point identified within the local point cloud reconstructed from real-time depth observations and $r_{safe}$ is the safety radius. Considering the second-order nature of quadrotor dynamics, we employ the HOCBF formulation to ensure the forward invariance of the safe set. The associated safety constraint is defined as:
\begin{equation}
	\ddot{B}_i(\mathbf{r}_t) + \alpha_1\dot{B}_i(\mathbf{r}_t) + \alpha_0 B_i(\mathbf{r}_t) \geq0,
\end{equation}
where $\alpha_0, \alpha_1 > 0$ are adjustable coefficients. 
In practice, these coefficients are empirically tuned to balance the trade-off between strict safety enforcement and trajectory reachability. Larger coefficients introduce excessive conservatism, while smaller coefficients lead to insufficient optimization effects. 
By substituting the time derivatives $\dot{B}_i(\textbf{r}_t) = 2 \mathbf{r}_t \cdot \mathbf{v}_t$ and $\ddot{B}_i(\textbf{r}_t) = 2 \| \mathbf{v}_t \|^2 + 2 \mathbf{r}_t \cdot \mathbf{a}_t$, we derive a linear inequality constraint on the control action $\mathbf{a}$:

\begin{equation}
	\underbrace{2 \mathbf{r}_t^{\top}}_{\mathbf{A}_{\mathrm{cbf}}}
	\mathbf{a}_t
	\ge
	\underbrace{-2\|\mathbf{v}_t\|^2 - \alpha_1 \dot{B}_i(\mathbf{r}_t) - \alpha_0 B_i(\mathbf{r}_t)}_{b_{\mathrm{cbf}}}.
\end{equation}
Here, $\mathbf{A}_{\text{cbf}}$ encapsulates the gradient direction for collision avoidance, while $b_{\text{cbf}}$ quantifies the minimum required control effort to counteract the inertia of the system and ensure boundary invariance. This linear constraint allows the QP solver to efficiently compute safe control commands $\mathbf{a}^*$ in real-time, effectively correcting unsafe maneuvers while preserving the intended flight trajectory. We evaluate the computational overhead of the HOCBF filter, and the result shows that the average optimization time for a single HOCBF execution is merely $90.61\,\mu\mathrm{s}$, with the maximum optimization time strictly bounded below $384.5\,\mu\mathrm{s}$. This ensures that the filter does not introduce any noticeable control latency during actual deployment.

\section{Experiments}
To validate the efficacy of the proposed framework, we first evaluate training performance across various reward configurations. Subsequently, ablation studies are conducted by deploying the policy in diverse, unseen environments to assess its impact on success rates. Furthermore, we benchmark the proposed method against state-of-the-art approaches, including the planning-based Ego-Planner~\cite{zhou2020ego} and the learning-based NavRL~\cite{xu2025navrl}, YOPO~\cite{yopo}, and DiffPhys~\cite{zhang2025learning}, demonstrating that our method consistently maintains the target velocity setpoint while ensuring rigorous safety. Real-world feasibility is verified through indoor experiments with varying obstacle configurations, compared against Ego-PlannerV2~\cite{zhou2022swarm}. Finally, we demonstrate the high-speed navigation capability of the framework in a cluttered outdoor forest, achieving agile flight at speeds up to 7.5 m/s.

\subsection{Simulations}

\subsubsection{Training Results}
Our framework is trained in Isaac Lab using large-scale parallel simulation, deploying 1,000 quadrotors simultaneously across 16 procedurally generated scenes of increasing difficulty. Each scene contains randomly placed trees and cylinders, with obstacle density gradually rising alongside the difficulty level. At the beginning of each episode, the initial position of the drone is uniformly randomized within the free space located before the obstacle field. An episode is considered successful if the drone reaches the goal region, defined as being within a 5-meter radius of the target position.

\begin{figure}[t]
	\centering
	\includegraphics[width=1.0\linewidth]{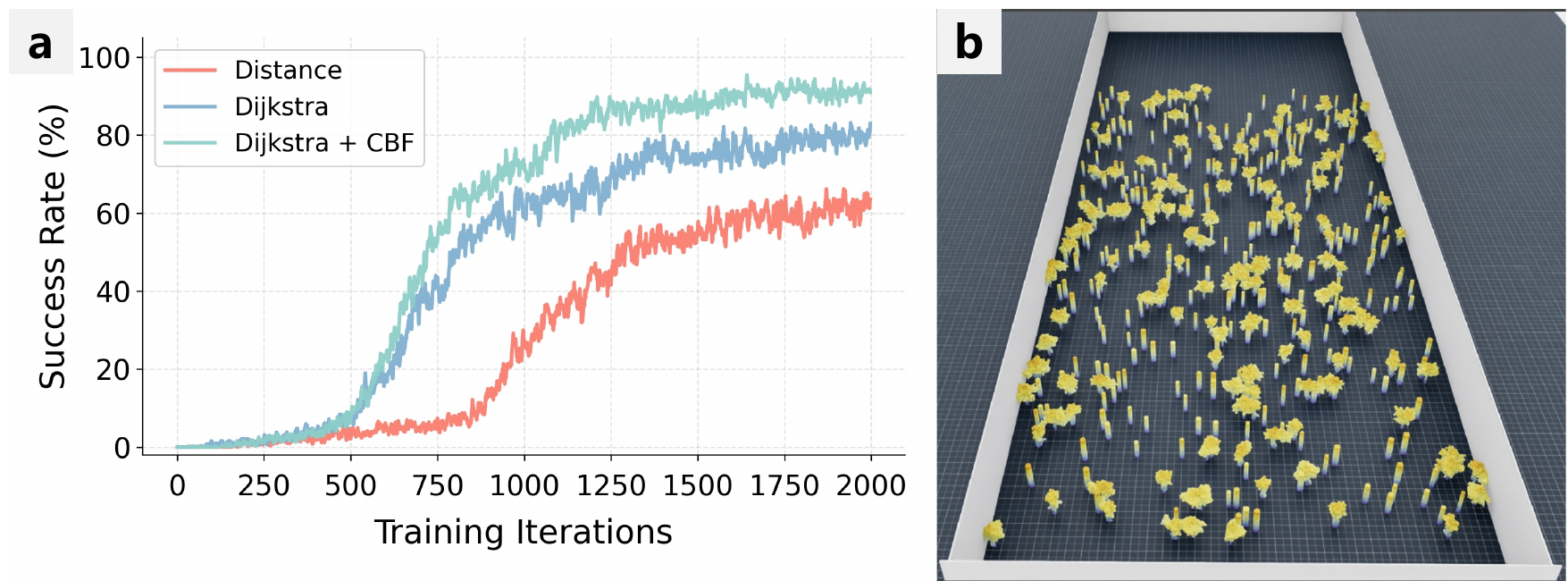} 
	\captionsetup{font={small}}
	\caption{\textbf{Training performance and environment.} (a) Comparative training curves displaying success rates over iterations for different reward configurations. (b) Visualization of a representative training scenario populated with dense obstacles.}
	\label{fig:training}
    \vspace{-8pt}
\end{figure}

To investigate the impact of reward shaping on learning efficiency and navigation performance, we compare three reward formulations: (i) Distance, which utilizes the Euclidean distance to the target as the navigation reward; (ii) Dijkstra, which replaces the Euclidean metric with a Dijkstra-based potential value for global guidance; and (iii) Dijkstra + CBF, which augments the Dijkstra reward with an additional CBF-based safety term to penalize proximity to obstacles. The learning curves in Fig.~\ref{fig:training}\hyperref[fig:training]{a} demonstrate that, under identical training conditions, the Dijkstra-based navigation reward substantially accelerates policy learning and achieves a consistently higher training success rate than the Distance baseline. Moreover, incorporating the CBF-based safety reward encourages safer exploration and yields further improvements in final performance, achieving the highest success rate among all compared variants.

\subsubsection{Ablation Studies}
\label{sec:ablation}
\begin{figure}[t]
	\centering
	\includegraphics[width=1.0\linewidth]{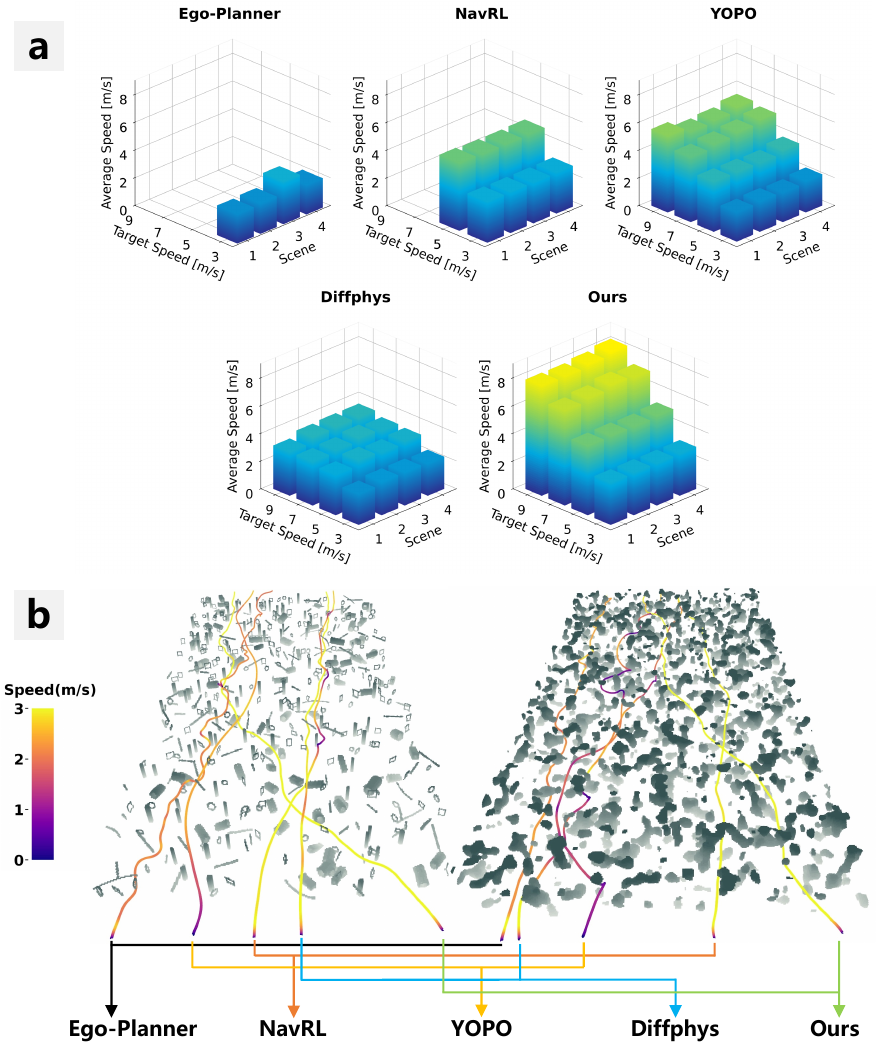}
	\captionsetup{font={small}}
	\caption{\textbf{Benchmark comparisons against state-of-the-art methods.} 
		\textbf{(a)} Average realized velocity over successful trials across varying target speeds on four representative scenes selected from the 16-scene benchmark, comparing Ego-Planner~\cite{zhou2020ego}, NavRL~\cite{xu2025navrl}, YOPO~\cite{yopo}, DiffPhys~\cite{zhang2025learning}, and our method. 
		\textbf{(b)} Visualization of the resulting flight trajectories generated by each method in geometric clutter (left) and Perlin noise (right) scenarios.}
	\label{fig:benchmark_traj}
	\vspace{-8pt}
\end{figure}

We assess the individual contributions of the Dijkstra-based navigation reward, the CBF-based safety reward, and the HOCBF safety filter through ablation experiments across eight unseen cluttered geometric scenes with dimensions of $64\times32\times3$ m. 
Unlike the training scenes, these scenes are composed of randomly placed inclined or vertical cylinders, frame structures, and thin poles that induce highly non-convex traversability (Fig.~\ref{fig:benchmark_traj}\hyperref[fig:benchmark_traj]{b}), while the deployed policy relies solely on its standard onboard observations without access to the privileged map information used during training for reward construction. We execute 50 independent trials per scene, where the start and goal positions are randomly sampled within rectangular regions of size $1\,\mathrm{m}\times32\,\mathrm{m}\times2\,\mathrm{m}$ at the starting line and the end of the environment, respectively, 
with the results summarized in Tab.~\ref{tab:ablation}. Overall, the results strongly validate our hierarchical design: global guidance ensures reachability, while the hybrid safety mechanism (soft shaping + hard filter) guarantees robustness across the entire speed envelope.

\textbf{Global guidance for reachability.} The Euclidean-distance baseline suffers rapid degradation at higher velocities due to its myopic nature, frequently leading to failure in complex clutter. In contrast, integrating the Dijkstra-based potential field significantly enhances reachability by encoding global topological information, which provides consistent gradients to steer the agent away from local minima (e.g., U-shaped traps). This strategy effectively distills long-horizon planning capabilities into the reactive policy without incurring online search latency. 
Notably, the slightly better performance at $5\,\mathrm{m/s}$ than at $3\,\mathrm{m/s}$ can be interpreted as a constraint-induced reachability phenomenon. With a stricter speed cap, the vehicle makes less forward progress per decision step and remains longer in the vicinity of non-convex obstacle structures, increasing its exposure to local trapping configurations. Moreover, because the nominal action is smaller in magnitude in the low-speed regime, the CBF-based correction constitutes a larger relative modification and can suppress the forward-progress component of the command. Consequently, failures at $3\,\mathrm{m/s}$ are more likely to arise from insufficient task progress and local stagnation than from loss of safety.

\textbf{Necessity of the hybrid safety architecture.}
Augmenting the policy with a CBF-based safety reward enhances performance in low-to-medium speed regimes. To isolate this contribution, we evaluated our formulation against a Geometric Inflation baseline across 160 trials in 16 unseen scenes at $5\,\mathrm{m/s}$, which enforced a doubled static safety margin of $0.4\,\mathrm{m}$ via terminal collision penalties. The inflation approach underperformed our CBF-guided policy, yielding both a lower success rate ($87.08\%$ vs. $92.50\%$) and a reduced average trajectory clearance ($0.89\,\mathrm{m}$ vs. $0.92\,\mathrm{m}$ ESDF). Simply enlarging static margins artificially obstructs narrow corridors, whereas our continuous CBF formulation provides velocity-aware, physics-grounded gradients that dynamically enhance safety without sacrificing reachability.
However, under high-dynamic conditions ($7\,\mathrm{m/s}$ and $9\,\mathrm{m/s}$), soft reward shaping exhibits a slight decline at high velocities, significant system inertia renders safety margins extremely sensitive to control inputs, making it challenging for the RL agent to approximate stiff constraints solely via reward feedback. This limitation underscores the critical necessity of our HOCBF safety filter. Acting as a deterministic safeguard during deployment, the filter solves a real-time QP to project raw commands onto a provably safe set. In complex simulation tests, the HOCBF filter was triggered in $13.06\%$ of the total control steps, successfully compensating for the neural policy's approximation errors in critical boundary cases. Consequently, the full framework effectively bridges the gap between learning-based agility and model-based safety.

\begin{table}[t]
	\vspace{5pt}
	\centering
	\captionsetup{font={small}}
	\caption{Results of Ablation Study Across Target Speeds}
	\label{tab:ablation}
	\setlength{\tabcolsep}{0pt} 
	\footnotesize 
	\begin{tabular*}{\columnwidth}{@{\extracolsep{\fill}} l cc cc cc cc }
		\toprule
		\multirow{2}{*}{\textbf{Configuration}} & \multicolumn{2}{c}{\textbf{3 m/s}} & \multicolumn{2}{c}{\textbf{5 m/s}} & \multicolumn{2}{c}{\textbf{7 m/s}} & \multicolumn{2}{c}{\textbf{9 m/s}} \\ 
		\cmidrule{2-3} \cmidrule{4-5} \cmidrule{6-7} \cmidrule{8-9}
		& SR$\uparrow$ & Vel & SR$\uparrow$ & Vel & SR$\uparrow$ & Vel & SR$\uparrow$ & Vel \\ \midrule
		Distance                & 51.75 & 3.04 & 35.00 & 5.16 & 21.25 & 7.11 & 0     & -    \\
		Dijkstra                & 62.25 & 3.02 & 67.75 & 5.18 & 56.25 & 6.76 & 43.75 & 8.36 \\
		Dijkstra + CBF          & 84.50 & 3.03 & 94.00 & 4.99 & 52.25 & 6.99 & 37.25 & 8.70 \\
		\textbf{Ours (Full)}    & \textbf{88.75} & 3.00 & \textbf{100} & 4.90 & \textbf{75.00} & 6.79 & \textbf{47.50} & 9.00 \\ \bottomrule
	\end{tabular*}
	\begin{flushleft}
		\scriptsize $^*$ SR: Success Rate (\%), Vel: Average Velocity (m/s).
	\end{flushleft}
	
\end{table}

\begin{table}[t]
	\centering
	\captionsetup{font={small}}
	\caption{Benchmark Comparison of Success Rates Across Target Speed}
	\label{tab:benchmark}
	\setlength{\tabcolsep}{0pt} 
	\footnotesize 
	\begin{tabular*}{\columnwidth}{@{\extracolsep{\fill}} l c c c c }
		\toprule
		\multicolumn{1}{c}{\textbf{Method}} & \textbf{3 m/s} & \textbf{5 m/s} & \textbf{7 m/s} & \textbf{9 m/s} \\ 
		\midrule
		Ego-Planner\cite{zhou2020ego} & 70.62\% & 0.62\% & 0.00\% & 0.00\% \\
		NavRL\cite{xu2025navrl} & 43.12\% & 5.00\% & 3.12\% & 0.00\% \\
		YOPO\cite{yopo} & 26.88\% & 21.25\% & 10.62\% & 6.88\% \\
		DiffPhys\cite{zhang2025learning} & 70.62\% & 55.62\% & 57.50\% & 51.25\% \\
		\textbf{Ours} & \textbf{96.88\%} & \textbf{96.25\%} & \textbf{73.75\%} & \textbf{56.88\%} \\ \bottomrule
	\end{tabular*}

\end{table}

\subsubsection{Benchmarks}
\label{sec:benchmarks}
We benchmark our proposed framework against four representative baselines: the planning-based Ego-Planner~\cite{zhou2020ego} and the learning-based NavRL~\cite{xu2025navrl}, YOPO~\cite{yopo}, and DiffPhys~\cite{zhang2025learning}. Each method was evaluated with its native observation design under the same target-speed budget. Experiments were conducted on 16 unseen benchmark scenes, all with the same dimensions as those used in Sec.~\ref{sec:ablation} and excluded from training (Fig.~\ref{fig:benchmark_traj}\hyperref[fig:benchmark_traj]{b}), categorized into two types: (1) 12 cluttered geometric scenes composed of randomly distributed primitives that induce highly non-convex traversability, and (2) 4 procedurally generated obstacle fields based on Perlin noise. Across these two scene types, scene difficulty increased with obstacle density, as reflected by the median free-space ESDF, which decreased from $1.24\,\mathrm{m}$ to $0.49\,\mathrm{m}$. For each method and target speed, we conducted 160 independent trials in total, and a run was counted as successful only if the UAV completed the scene without collision and finished within $5\,\mathrm{m}$ of the goal. Representative success-rate results are reported in Tab.~\ref{tab:benchmark}.

\begin{figure}[t]
	\centering
	\captionsetup[subfigure]{font={small},skip=2pt}
	
	\begin{subfigure}[b]{\linewidth}
		\centering
		\includegraphics[width=1.0\linewidth]{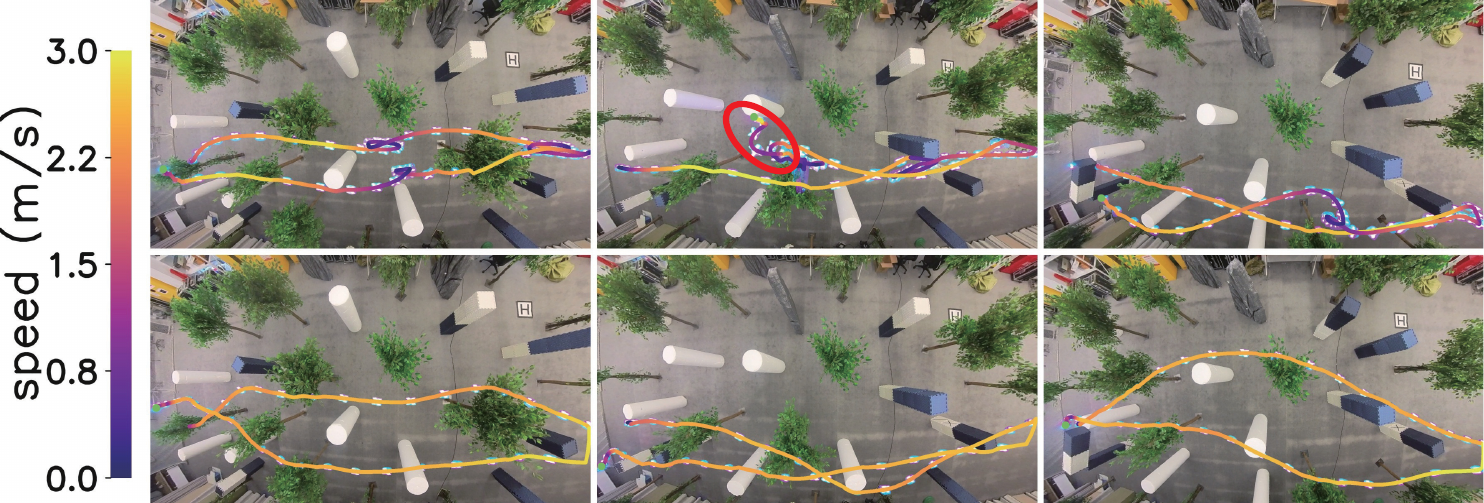}
		\caption{Target speed: 3 m/s}
		\label{fig:indoor_3ms}
	\end{subfigure}
	
	\vspace{1.5pt} 
	
	\begin{subfigure}[b]{\linewidth}
		\centering
		\includegraphics[width=1.0\linewidth]{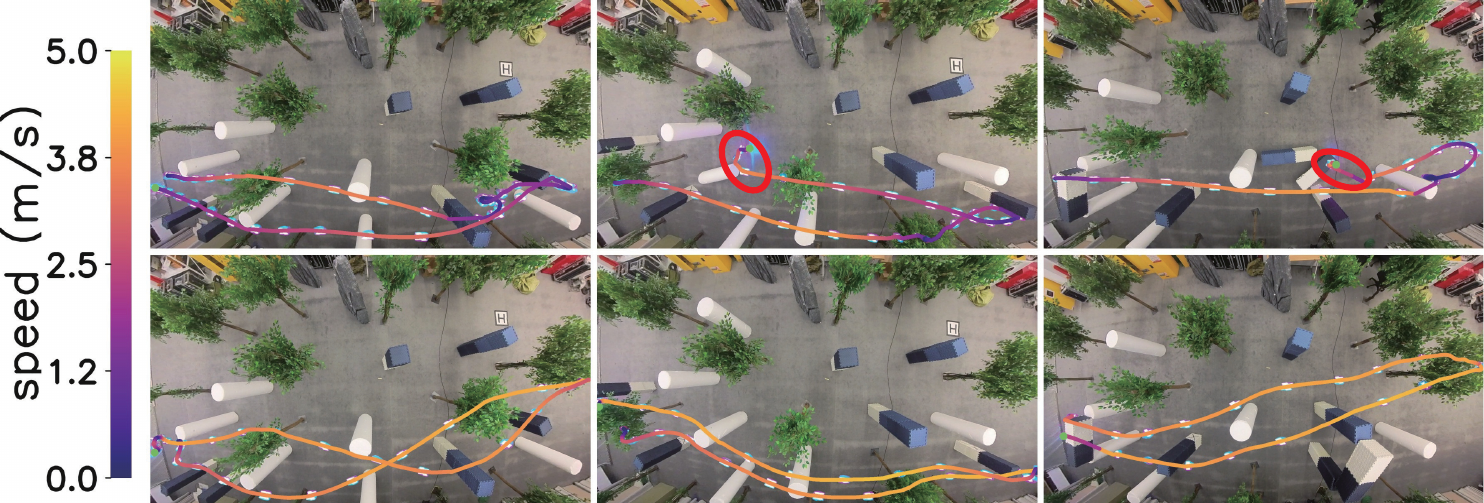}
		\caption{Target speed: 5 m/s}
		\label{fig:indoor_5ms}
	\end{subfigure}
	
	\vspace{1.5pt} 
	
	\begin{subfigure}[b]{\linewidth}
		\centering
		\includegraphics[width=1.0\linewidth]{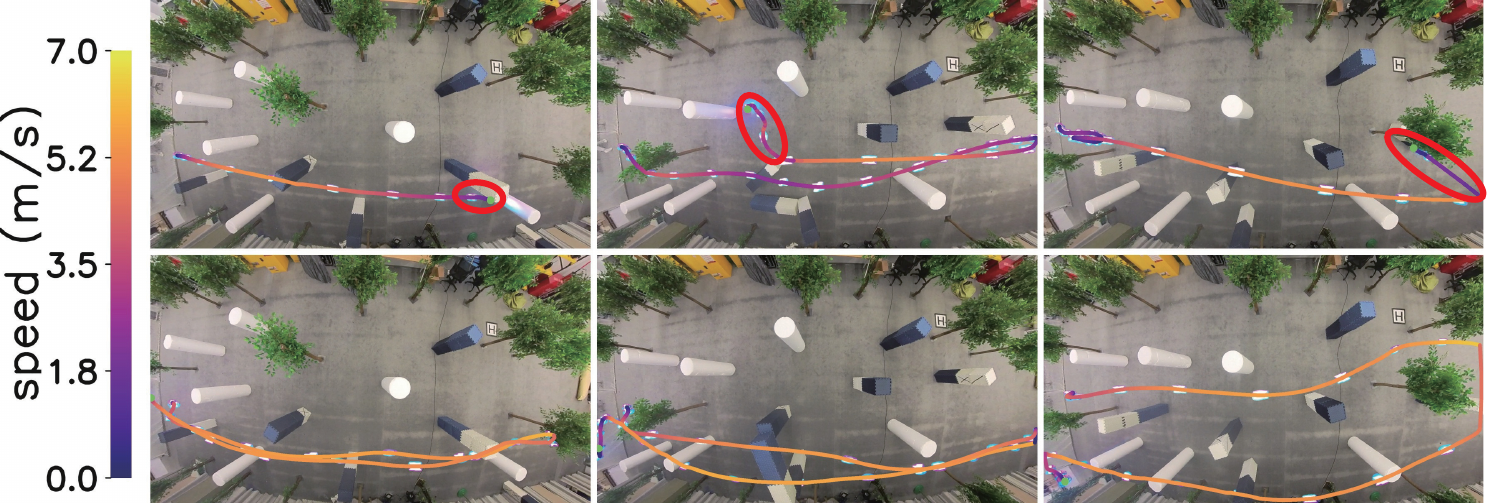}
		\caption{Target speed: 7 m/s}
		\label{fig:indoor_7ms}
	\end{subfigure}
	
	\captionsetup{font={small}}
	\caption{Snapshots of indoor real-world experiments at varying target speeds. 
		Each subfigure demonstrates three independent trials conducted under distinct obstacle configurations to evaluate robustness. 
		In each panel, the top and bottom rows compare the performance of Ego-Planner \textbf{(Top)} and our proposed method \textbf{(Bottom)}, respectively.}
	\label{fig:indoor_experiments}
	\vspace{-10pt} 
\end{figure}

As detailed in Tab.~\ref{tab:benchmark}, Ego-Planner~\cite{zhou2020ego} remains viable only at the lowest target speed and fails almost completely beyond $3\,\mathrm{m/s}$, reflecting the latency bottleneck of iterative mapping and replanning in dense clutter. Among the learning-based baselines, NavRL~\cite{xu2025navrl} degrades most sharply with speed, dropping from $43.12\%$ at $3\,\mathrm{m/s}$ to nearly zero thereafter. This behavior is likely attributable to the mismatch between its relatively regularized training scenes and the highly cluttered benchmark environments, together with its static-obstacle input design, which limits robustness as the reaction time shrinks at higher speed. YOPO~\cite{yopo} is more stable than NavRL and can track the target speed reasonably well, but its depth-conditioned selection over a finite set of short-horizon motion primitives remains vulnerable in highly non-convex clutter, where a locally favorable primitive can still steer the vehicle into dead ends or collisions. DiffPhys~\cite{zhang2025learning} achieves higher success rates than NavRL and YOPO, but its realized speed saturates near $3\,\mathrm{m/s}$ (Fig.~\ref{fig:benchmark_traj}\hyperref[fig:benchmark_traj]{a}). This limited high-speed scalability likely stems from tying the camera heading to commanded motion. During aggressive maneuvers, large pitch excursions shift the camera view towards the ground or sky, severely reducing forward obstacle visibility. Consequently, the policy adopts conservative behavior, trading speed for perceptual stability and failing at higher target speeds.

Conversely, our method achieves the optimal safety-agility trade-off, maintaining the highest success rates and realized velocities across all speeds and configurations (Fig.~\ref{fig:benchmark_traj}\hyperref[fig:benchmark_traj]{a}). Specifically, the success rate remains at $96.88\%$, $96.25\%$, $73.75\%$, and $56.88\%$ from $3$ to $9\,\mathrm{m/s}$, with the average successful speed increasing from $3.00$ to $7.58\,\mathrm{m/s}$. These results validate the advantage of combining low-latency learning with model-based safety: the learned policy preserves aggressive speed tracking, while the safety layer enforces reliable near-obstacle behavior and improves robustness as the environment and target speed become increasingly challenging.

\subsection{Real-World Experiments}

To further validate the proposed framework, we deployed the trained policy on a physical quadrotor and conducted extensive experiments in both cluttered indoor environments and unstructured outdoor forests.


We construct a cluttered indoor environment, as shown in Fig.~\ref{fig:indoor_experiments}. The policy is evaluated using networks trained with target speed constraints of $3 \text{ m/s}$, $5 \text{ m/s}$, and $7 \text{ m/s}$. To assess generalization capabilities, the quadrotor is tasked with navigating back and forth through a $15 \text{ m}$ long obstacle course, with layouts randomly reconfigured before each flight.

Our policy achieves exceptionally high success rates in obstacle avoidance and navigation across diverse speeds and configurations. Compared to the traditional Ego-PlannerV2~\cite{zhou2022swarm}, it demonstrates superior flight stability and maximum achievable velocity. As shown in Fig.~\ref{fig:indoor_experiments} (where red circles denote collisions), Ego-PlannerV2~\cite{zhou2022swarm} failures increase significantly at higher target speeds. In contrast, our method consistently achieves successful navigation while maintaining an actual velocity that closely tracks the target setpoint for the majority of the trajectory.


\begin{figure}[t]
	\centering
	\includegraphics[width=0.8\linewidth]{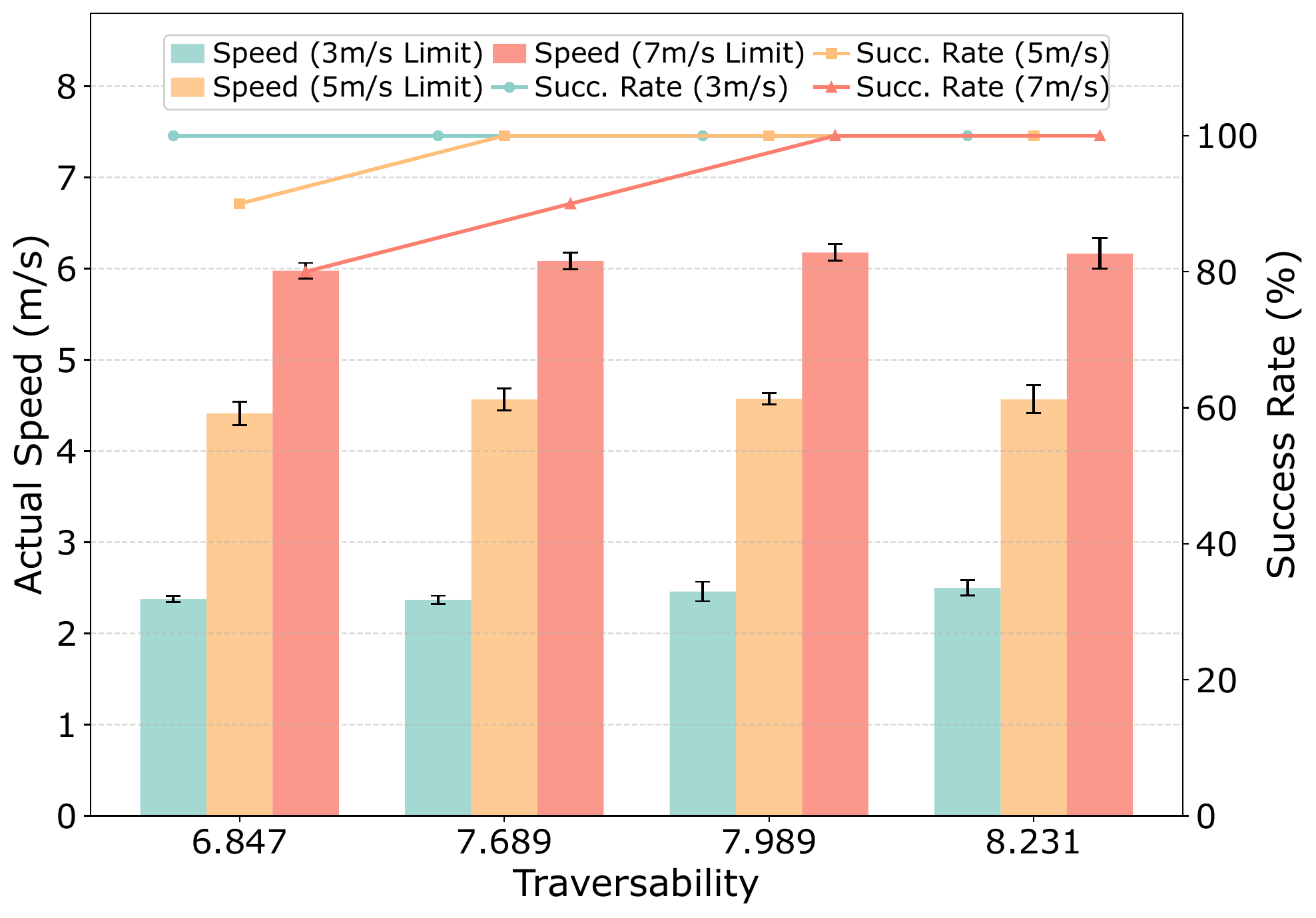}
	\captionsetup{font={small}}
    \caption{Quantitative evaluation of flight performance relative to environment complexity. The figure illustrates the success rates and statistical velocity measures (mean and standard deviation) under varying target speeds across four indoor environments with decreasing traversability.}
	\label{fig:success}
	\vspace{-8pt}
\end{figure}

To provide a more principled evaluation of real-world scene complexity, we further quantify the difficulty of the indoor test environments using the traversability metric following \cite{yu2023avoidbench}. Four indoor environments with increasing clutter levels are constructed, and the policy is evaluated under three target-speed settings, namely 3 m/s, 5 m/s, and 7 m/s. For each combination of environment complexity and target speed, 10 independent real-world trials are conducted.

As shown in Fig.~\ref{fig:success}, the proposed method achieves a 100\% success rate for all target speeds in the two lower-difficulty environments. In more cluttered scenes, the policy remains robust: the 5 m/s setting fails only once in the most difficult environment, while the 7 m/s setting achieves success rates of 90\% and 80\% in the two most challenging environments, respectively. The corresponding velocity statistics further show that the realized speed remains close to the target speed and does not exhibit a significant decrease as the environment complexity increases. These results indicate that the proposed policy maintains both safety and agility under increasing clutter, rather than relying on conservative speed reduction.

\begin{figure}[t]
	\centering
	\includegraphics[width=1.0\linewidth]{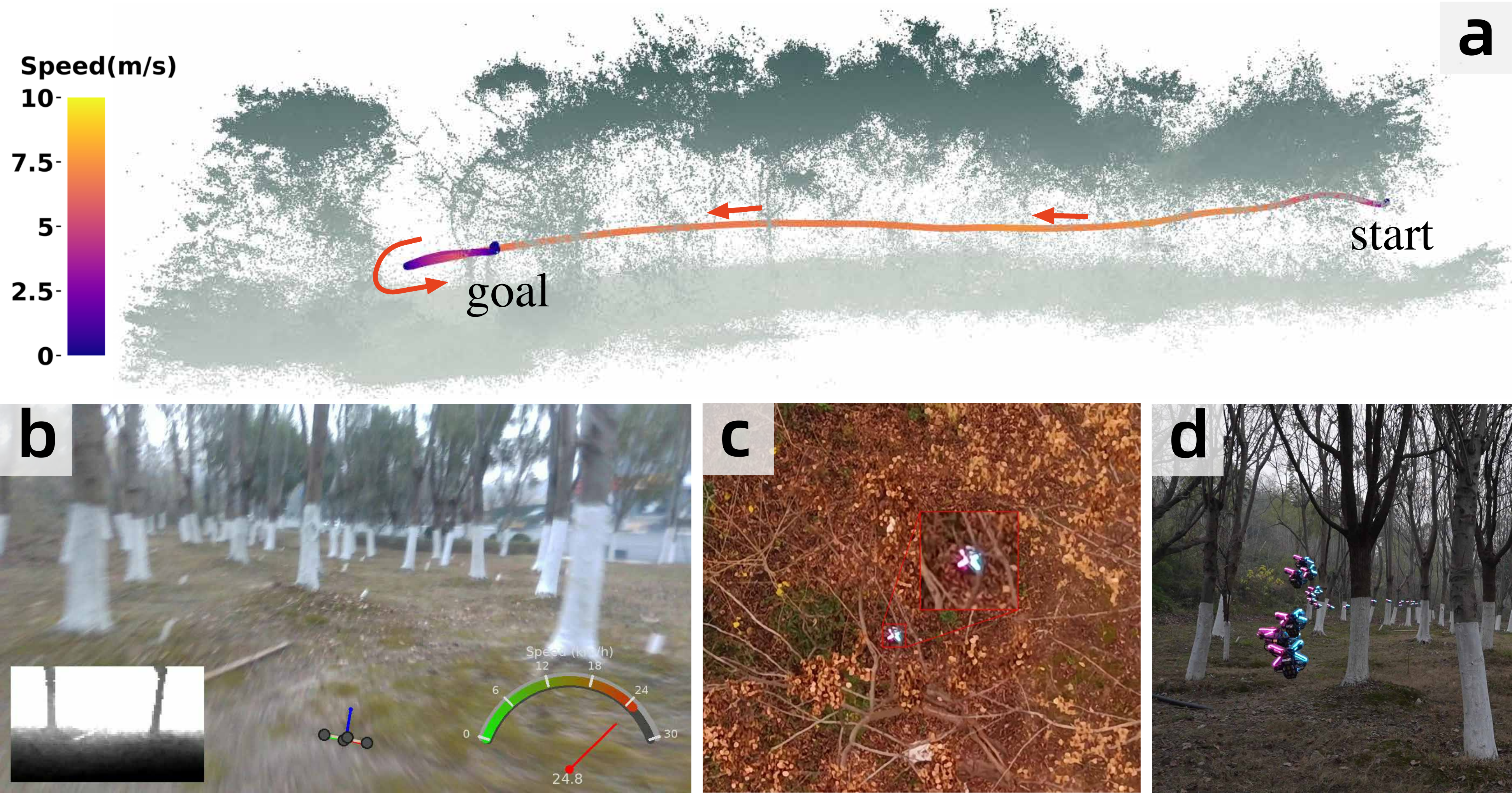}
	\captionsetup{font={small}}
	\caption{\textbf{Visualization of the outdoor flight experiment.} \textbf{(a)} Global point cloud map overlaid with the executed trajectory and the corresponding realized speed profile. \textbf{(b)} Egocentric observations, including RGB and aligned depth images (Intel RealSense D435i), augmented with real-time attitude and speedometer displays. \textbf{(c)} Top-down view of the environment. \textbf{(d)} Side-view composite image illustrating the sequential flight maneuvers through the dense vegetation.}
	\label{fig:outdoor}
	\vspace{-8pt}
\end{figure}

In outdoor forest scenarios (Fig.~\ref{fig:outdoor}), we employ a policy trained with a setup modified only by reducing obstacle density and increasing the target speed limit compared to the original training configuration. The quadrotor successfully achieves stable, high-speed, long-distance obstacle avoidance, maintaining an average velocity of $7 \text{ m/s}$ over a distance exceeding $35 \text{ m}$ to reach the target.

Furthermore, we evaluated our policy under dynamic conditions in simulation and the real world. Despite being trained solely in static environments, the UAV successfully generalized to unknown dynamic scenarios, achieving a $64\%$ success rate across 100 simulated trials with dense moving obstacles, and completing all 10 real-world flight tests without collisions against an obstacle moving at $0.5 \text{ m/s}$. 
The failures primarily occur when moving obstacles approach from the sensor's blind spots, rendering the local perception horizon insufficient for successful reactive maneuvers.

\section{Conclusion}
In this paper, we present a hybrid reinforcement learning framework integrating model-free policy optimization with model-based physical priors for high-speed quadrotor obstacle avoidance. The priors accelerate training convergence via geometric guidance and ensure feasible control deployment through real-time filtering. 
Currently, hardware limitations restrict real-world flight speeds from reaching the policy's maximum simulated capabilities. Future work will focus on enhancing quadrotor hardware performance and extending this method to specific, task-oriented scenarios.

\bibliographystyle{IEEEtran}
\bibliography{ref}

@string{icra = {Proc. of the {IEEE} Intl. Conf. on Robot. and Autom.}}

@article{ppo,
  author       = {John Schulman and
                  Filip Wolski and
                  Prafulla Dhariwal and
                  Alec Radford and
                  Oleg Klimov},
  title        = {Proximal Policy Optimization Algorithms},
  journal      = {CoRR},
  volume       = {abs/1707.06347},
  year         = {2017},
  url          = {http://arxiv.org/abs/1707.06347},
  eprinttype    = {arXiv},
  eprint       = {1707.06347},
  bibsource    = {dblp computer science bibliography, https://dblp.org}
}

@article{loquercio2021learning,
  title={Learning high-speed flight in the wild},
  author={Loquercio, Antonio and Kaufmann, Elia and Ranftl, Ren{\'e} and M{\"u}ller, Matthias and Koltun, Vladlen and Scaramuzza, Davide},
  journal={Science Robotics},
  volume={6},
  number={59},
  pages={eabg5810},
  year={2021},
  publisher={American Association for the Advancement of Science}
}

@article{song2023reaching,
  title={Reaching the limit in autonomous racing: Optimal control versus reinforcement learning},
  author={Song, Yunlong and Romero, Angel and M{\"u}ller, Matthias and Koltun, Vladlen and Scaramuzza, Davide},
  journal={Science Robotics},
  volume={8},
  number={82},
  pages={eadg1462},
  year={2023},
  publisher={American Association for the Advancement of Science}
}

@article{wu2024whole,
  title={Whole-Body Control Through Narrow Gaps From Pixels To Action},
  author={Wu, Tianyue and Chen, Yeke and Chen, Tianyang and Zhao, Guangyu and Gao, Fei},
  journal={arXiv preprint arXiv:2409.00895},
  year={2024}
}

@article{zhou2020ego,
  title={Ego-planner: An esdf-free gradient-based local planner for quadrotors},
  author={Zhou, Xin and Wang, Zhepei and Ye, Hongkai and Xu, Chao and Gao, Fei},
  journal={IEEE Robotics and Automation Letters},
  volume={6},
  number={2},
  pages={478--485},
  year={2020},
  publisher={IEEE}
}

@article{zhou2019robust,
  title={Robust and efficient quadrotor trajectory generation for fast autonomous flight},
  author={Zhou, Boyu and Gao, Fei and Wang, Luqi and Liu, Chuhao and Shen, Shaojie},
  journal={IEEE Robotics and Automation Letters},
  volume={4},
  number={4},
  pages={3529--3536},
  year={2019},
  publisher={IEEE}
}

@article{gao2020teach,
  title={Teach-repeat-replan: A complete and robust system for aggressive flight in complex environments},
  author={Gao, Fei and Wang, Luqi and Zhou, Boyu and Zhou, Xin and Pan, Jie and Shen, Shaojie},
  journal={IEEE Transactions on Robotics},
  volume={36},
  number={5},
  pages={1526--1545},
  year={2020},
  publisher={IEEE}
}

@article{zhou2022swarm,
  title={Swarm of micro flying robots in the wild},
  author={Zhou, Xin and Wen, Xiangyong and Wang, Zhepei and Gao, Yuman and Li, Haojia and Wang, Qianhao and Yang, Tiankai and Lu, Haojian and Cao, Yanjun and Xu, Chao and others},
  journal={Science Robotics},
  volume={7},
  number={66},
  pages={eabm5954},
  year={2022},
  publisher={American Association for the Advancement of Science}
}

@ARTICLE{yopo,
  author={Lu, Junjie and Zhang, Xuewei and Shen, Hongming and Xu, Liwen and Tian, Bailing},
  journal={IEEE Robotics and Automation Letters}, 
  title={You Only Plan Once: A Learning-Based One-Stage Planner With Guidance Learning}, 
  year={2024},
  volume={9},
  number={7},
  pages={6083-6090},
  doi={10.1109/LRA.2024.3399589}}

@article{yu2024mavrl,
	title={MAVRL: Learn to fly in cluttered environments with varying speed},
	author={Yu, Hang and De Wagter, Christophe and de Croon, Guido CH E},
	journal={IEEE Robotics and Automation Letters},
	year={2024},
	publisher={IEEE}
}

@article{zhang2025learning,
	title={Learning vision-based agile flight via differentiable physics},
	author={Zhang, Yuang and Hu, Yu and Song, Yunlong and Zou, Danping and Lin, Weiyao},
	journal={Nature Machine Intelligence},
	pages={1--13},
	year={2025},
	publisher={Nature Publishing Group UK London}
}

@article{xu2025navrl,
	title={Navrl: Learning safe flight in dynamic environments},
	author={Xu, Zhefan and Han, Xinming and Shen, Haoyu and Jin, Hanyu and Shimada, Kenji},
	journal={IEEE Robotics and Automation Letters},
	year={2025},
	publisher={IEEE}
}

@article{ames2016control,
	title={Control barrier function based quadratic programs for safety critical systems},
	author={Ames, Aaron D and Xu, Xiangru and Grizzle, Jessy W and Tabuada, Paulo},
	journal={IEEE Transactions on Automatic Control},
	volume={62},
	number={8},
	pages={3861--3876},
	year={2016},
	publisher={IEEE}
}

@inproceedings{ames2019control,
	title={Control barrier functions: Theory and applications},
	author={Ames, Aaron D and Coogan, Samuel and Egerstedt, Magnus and Notomista, Gennaro and Sreenath, Koushil and Tabuada, Paulo},
	booktitle={2019 18th European control conference (ECC)},
	pages={3420--3431},
	year={2019},
	organization={Ieee}
}

@article{xiao2021hocbf,
	title={High-order control barrier functions},
	author={Xiao and Belta},
	journal={IEEE Transactions on Automatic Control},
	volume  = {67},
	number  = {7},
	pages={3655--3662},
	year={2021},
	organization={IEEE}
}

@article{Zhou2021FUEL,
  author    = {Bo Zhou and Yifan Zhang and Xin Chen and Fei Gao},
  title     = {{FUEL}: Fast UAV Exploration Using Incremental Frontier Structure and Hierarchical Planning},
  journal   = {IEEE Robotics and Automation Letters},
  volume    = {6},
  number    = {2},
  pages     = {779--786},
  year      = {2021},
  doi       = {10.1109/LRA.2020.3047779}
}

@article{scaramuzza2014vision,
  title={Vision-controlled micro flying robots: from system design to autonomous navigation and mapping in GPS-denied environments},
  author={Scaramuzza, Davide and Achtelik, Michael C and Doitsidis, Lefteris and Friedrich, Fraundorfer and Kosmatopoulos, Elias and Martinelli, Agostino and Achtelik, Markus W and Chli, Margarita and Chatzichristofis, Savvas and Kneip, Laurent and others},
  journal={IEEE Robotics \& Automation Magazine},
  volume={21},
  number={3},
  pages={26--40},
  year={2014},
  publisher={IEEE}
}

@article{xu2025flying,
  title={Flying on Point Clouds with Reinforcement Learning},
  author={Xu, Guangtong and Wu, Tianyue and Wang, Zihan and Wang, Qianhao and Gao, Fei},
  journal={arXiv preprint arXiv:2503.00496},
  year={2025}
}

@article{han2025reactive,
  title={Reactive Aerobatic Flight via Reinforcement Learning},
  author={Han, Zhichao and Huang, Xijie and Xu, Zhuxiu and Zhang, Jiarui and Wu, Yuze and Wang, Mingyang and Wu, Tianyue and Gao, Fei},
  journal={arXiv preprint arXiv:2505.24396},
  year={2025}
}

@article{han2025dynamically,
  title={Dynamically feasible trajectory generation with optimization-embedded networks for autonomous flight},
  author={Han, Zhichao and Xu, Long and Pei, Liuao and Gao, Fei},
  journal={IEEE Robotics and Automation Letters},
  year={2025},
  publisher={IEEE}
}

@article{xu2023vision,
  title={A vision-based autonomous UAV inspection framework for unknown tunnel construction sites with dynamic obstacles},
  author={Xu, Zhefan and Chen, Baihan and Zhan, Xiaoyang and Xiu, Yumeng and Suzuki, Christopher and Shimada, Kenji},
  journal={IEEE Robotics and Automation Letters},
  volume={8},
  number={8},
  pages={4983--4990},
  year={2023},
  publisher={IEEE}
}

@inproceedings{nguyen2022motion,
  title={Motion primitives-based navigation planning using deep collision prediction},
  author={Nguyen, Huan and Fyhn, Sondre Holm and De Petris, Paolo and Alexis, Kostas},
  booktitle={2022 International Conference on Robotics and Automation (ICRA)},
  pages={9660--9667},
  year={2022},
  organization={IEEE}
}

@article{kahn2021badgr,
  title={Badgr: An autonomous self-supervised learning-based navigation system},
  author={Kahn, Gregory and Abbeel, Pieter and Levine, Sergey},
  journal={IEEE Robotics and Automation Letters},
  volume={6},
  number={2},
  pages={1312--1319},
  year={2021},
  publisher={IEEE}
}

@article{song2022learning,
  title={Learning perception-aware agile flight in cluttered environments},
  author={Song, Yunlong and Shi, Kexin and Penicka, Robert and Scaramuzza, Davide},
  journal={arXiv preprint arXiv:2210.01841},
  year={2022}
}

@ARTICLE{wu2024clutter,
  author={Zhao, Guangyu and Wu, Tianyue and Chen, Yeke and Gao, Fei},
  journal={IEEE Robotics and Automation Letters}, 
  title={Learning Speed Adaptation for Flight in Clutter}, 
  year={2024},
  volume={9},
  number={8},
  pages={7222-7229},
  doi={10.1109/LRA.2024.3421789}}

@ARTICLE{Liu2017sfc,
  author={Liu, Sikang and Watterson, Michael and Mohta, Kartik and Sun, Ke and Bhattacharya, Subhrajit and Taylor, Camillo J. and Kumar, Vijay},
  journal={IEEE Robotics and Automation Letters}, 
  title={Planning Dynamically Feasible Trajectories for Quadrotors Using Safe Flight Corridors in 3-D Complex Environments}, 
  year={2017},
  volume={2},
  number={3},
  pages={1688-1695},
  doi={10.1109/LRA.2017.2663526}}

@article{emam2022safe,
  title={Safe reinforcement learning using robust control barrier functions},
  author={Emam, Yousef and Notomista, Gennaro and Glotfelter, Paul and Kira, Zsolt and Egerstedt, Magnus},
  journal={IEEE Robotics and Automation Letters},
  volume={10},
  number={3},
  pages={2886--2893},
  year={2022},
  publisher={IEEE}
}

@inproceedings{cheng2019end,
  title={End-to-end safe reinforcement learning through barrier functions for safety-critical continuous control tasks},
  author={Cheng, Richard and Orosz, G{\'a}bor and Murray, Richard M and Burdick, Joel W},
  booktitle={Proceedings of the AAAI conference on artificial intelligence},
  volume={33},
  number={01},
  pages={3387--3395},
  year={2019}
}

@article{lee2025quadrotor,
  title={Quadrotor Navigation using Reinforcement Learning with Privileged Information},
  author={Lee, Jonathan and Rathod, Abhishek and Goel, Kshitij and Stecklein, John and Tabib, Wennie},
  journal={arXiv preprint arXiv:2509.08177},
  year={2025}
}

@article{wu2026precise,
  title={Precise aggressive aerial maneuvers with sensorimotor policies},
  author={Wu, Tianyue and Xu, Guangtong and Wang, Zihan and Lin, Junxiao and Chen, Tianyang and Wu, Yuze and Han, Zhichao and Liu, Zhiyang and Gao, Fei},
  journal={Science Robotics},
  volume={11},
  number={115},
  pages={eaeb0180},
  year={2026},
  publisher={American Association for the Advancement of Science}
}

@article{yu2023avoidbench,
  title={Avoidbench: A high-fidelity vision-based obstacle avoidance benchmarking suite for multi-rotors},
  author={Yu, Hang and de Croon, Guido CH and De Wagter, Christophe},
  journal={arXiv preprint arXiv:2301.07430},
  year={2023}
}

\vfill

\end{document}